\documentclass[11pt]{article} 

\usepackage{graphicx}
\usepackage{booktabs}
\usepackage{algorithm}
\usepackage{algpseudocode}
\usepackage[T1]{fontenc}
\usepackage{lmodern}
\usepackage[flushleft]{threeparttable}
\usepackage[hidelinks]{hyperref} %
\hypersetup{hidelinks}
\usepackage{multirow}
\usepackage{longtable}
\usepackage[margin=3cm]{geometry} %
\usepackage{amsmath,amsthm,amssymb}
\usepackage{authblk}
\usepackage{caption}
\usepackage{microtype}

\title{Latent space configuration for neural network finetuning, fast
ranking, and embedding dimensionality reduction}
\author[1]{Nikita Gabdullin}
\author[1]{Anton Raskovalov}
\affil[1]{Joint Stock "Research and production company "Kryptonite" \authorcr
E-mail: n.gabdullin@kryptonite.ru, a.raskovalov@kryptonite.ru}
\date{}
\begin{document}

    \captionsetup[table]{labelformat={default},labelsep=period,name={Table}}

    \maketitle
    \begin{abstract}
        Finding \emph{K} most similar elements for a query in
        a large gallery of neural network (NN) embeddings is essential in
        various areas of computer science and industrial applications of
        artificial intelligence. This paper proposes search algorithms for
        structured latent spaces (LS) of NNs which allow to speed up top
        \emph{K} item retrieval owing to $\mathcal{O}(1)$ complexity of a small candidate
        gallery subset search routine. ArcFace-pretrained IResNet50 models are finetuned
        using Latent Space Configuration (LSC) methodology to obtain
        new distribution of embeddings with desired geometric
        properties. The application of the proposed algorithms to embeddings obtained from
        finetuned models leads to several times speed-up over baseline
        exhaustive search. Specifically, we report up to 3 times speed-up on GPU
        and 40\% speed-up on CPU compared to the baseline method GPU speed.
        These results are obtained for the original and reduced numbers of LS
        dimensions. The results show that LS dimensionality reduction allows to
        further accelerate the search and even improve rank-\emph{N} metrics.
  
    \end{abstract}

    
    \section{Introduction}
\label{introduction}

Nowadays, various complex systems rely on efficient retrieval of
relevant information from extremely large databases. Often this requires
data ranking by asserting relevance of database data with the query and
returning \emph{K} most relevant elements (top-\emph{K}, or
top-\emph{k}, or top-\emph{N})~\cite{topksurv}. High-dimensional outputs
(embeddings) of neural networks (NNs) can directly be used as
description vectors that represent the data, so numerical similarity
between vectors becomes a measure of relevance~\cite{BSG}. This is widely
used in document analysis, recommendation systems, speech and audio
recognition~\cite{NCF, EC, string}. In context of computer vision (CV),
this translates to finding \emph{K} most relevant images in gallery
\emph{G} for a query \emph{q}. This is essential for e-commerce, face
recognition, biometrics, and security~\cite{idsurve, BSG, CLIP}.

Exhaustive search has $\mathcal{O}(n\textsubscript{g})$ complexity since it
requires to compare queries with all embeddings in \emph{G} with size
\emph{n\textsubscript{g}}. This is not feasible for galleries of million
or billions of elements, so approximate methods are used instead. These
methods often require building indexing structures over the gallery for
fast retrieval of a smaller subset specific to the given query~\cite{HNSW,
PQ, LSH}. While indeed allowing sub-linear scaling, these methods are 
still computationally expensive and can reduce complexity to
$\mathcal{O}(\sqrt{n\textsubscript{g}})$~\cite{VDB}. 

In this paper we take advantage of geometric properties of vector
systems used for NN latent space configuration (LSC) to find a small
subset in \emph{G} and significantly accelerate top \emph{K} retrieval.
LSC is a novel methodology which uses predefined class prototypes to
structure and add geometric properties to NN LS during training
~\cite{lsc2}. It has previously been shown that the complexity of the
search of mean vectors of class embeddings for classifier NNs with
preconfigured LS is $\mathcal{O}(1)$, which can be realized with a single
sort operation performed on the embedding vector. In context of face
recognition this translates to the search of mean vectors of person
identity (\emph{id}) embedding clusters with $\mathcal{O}(1)$ complexity.
Hence, the complete search complexity becomes $\mathcal{O}(n\textsubscript{c})$
where \emph{n\textsubscript{c}} is
\emph{id} cluster size which can be several orders of magnitude smaller
than \emph{n\textsubscript{g}}. It is also important that
\emph{n\textsubscript{c}} depends on the number of images per \emph{id}
(which can differ for different \emph{id}s) and not
\emph{n\textsubscript{g}}.

The desired geometry of NN emebdding distribution can be achieved with
finetuning. NN pretraining on extremely large general purpose datasets
and finetuning on smaller target datasets using supervised or
self-supervied learning methods has proven to be very effective in
various CV areas~\cite{BiT, CLIP, VIT}. It has recently been shown that
finetuning can also be performed using prototype-based training methods~\cite{PCL, PCL-CLIP}. 
However, most of them use trainable prototypes and
require loss functions which complexity grows linearly with the number
of clusters, classes, or \emph{id}s in the gallery
\emph{n\textsubscript{id }}~\cite{proto, PCL, CenterLoss}. On the
contrary, LSC chooses predefined non-trainable prototypes allowing to
avoid \emph{n\textsubscript{id}} scaling, which is beneficial for
training and finetuning on datasets with extremely large
\emph{n\textsubscript{id}}~\cite{lsc2}.

In this paper we propose two \emph{id} cluster search algorithms which
allow to rapidly identify a small candidate subset of gallery embeddings
which similarity with the query needs to be assessed. These algorithms
require that NN LS possesses certain geometric properties, so
configuring LS during finetuning is essential for fast high-accuracy
retrieval. The application of the proposed algorithms to embeddings of
finetuned NN leads to several times speed-up on GPU and 40\% speed-up on
CPU compared to exhaustive search performed on GPU. The latter is
particularly important for low computational budget and edge
applications where GPU resources are limited or unavailable.

The rest of the paper is organized as follows: Section~\ref{lscfinetune} outlines the
LSC methodology for finetuning, Section~\ref{search} describes the proposed search
algorithms, Section~\ref{Exp} summarizes experimental results, Section~\ref{discussions}
discusses potential application scenarios of proposed methods, and
Section~\ref{conclusions} concludes the paper.

\section{Configuring LS of NNs during training and finetuning}
\label{lscfinetune}

LSC training methodology is discussed in detail in~\cite{lsc2, lsc1}. The
goal of LSC is to force mean vectors of embedding clusters to align with
vectors of some predefined vector system \emph{V\textsubscript{n}} with
desired geometric properties. To configure NN LS one needs to choose the
base \emph{V\textsubscript{n}}, to distribute mean vectors of embedding
clusters (corresponding to classes or \emph{ids}) among them, and to
choose a loss function. It has been shown that cosine loss (which
maximizes cosine similarity between embeddings and prototype vectors)
performs best in high-dimensional cases~\cite{lsc2}. Since cosine loss
values illustrate how close \emph{id} embeddings are to their target
vectors, training naturally reduces cluster size while ensuring
separation of clusters determined by the \emph{V\textsubscript{n}}
vector distribution.

It has been shown that NN discriminative ability depends on properties
of the LS embedding distribution. Firstly, embeddings of similar inputs must
have high similarity and be close in LS, so clusters of embeddings
corresponding to one class or \emph{id} are compact. Secondly,
embeddings of different classes or \emph{ids} must have low similarity,
so embedding clusters are well-separated. This is achieved by
introducing loss functions that affect NN embedding distribution by
acting directly in LS~\cite{CenterLoss, SphereFace}. One of the most
important methods is ArcFace which has introduced an additive angular
margine penalty to standard softmax loss to improve cluster
compactization and separation~\cite{ArcFace}. This has led to significant
improvement in discriminative ability and the overall performance of
face recognition NNs.

However, ArcFace requires to use several complicated loss functions
which include \emph{id} loss, which requires having an NN layer with
\emph{n\textsubscript{id}} size. As previously mentioned, this limits
the application of this method when \emph{n\textsubscript{id}} is
extremely large. On the contrary, LSC does not require creating
trainable objects of \emph{n\textsubscript{id}} size and, by choosing
sparsely distributed prototypes, the inter- and intra-class (cluster)
distribution requirements mentioned above are automatically satisfied
when clusters formed around \emph{V\textsubscript{n}} vectors are
sufficiently compact.

In general, any system of vectors with some predefined set of desired
properties can be used for LSC. In this paper we use systems of vectors
\(V_{n}^{mk}\) such that all vectors in the system are obtained as
permutations of elements of a base vector which consists of \emph{m}
ones, \emph{k} negative ones, and (\(n - m - k)\) zeros~\cite{lsc3,
lsc4}. Specifically, we use \(V_{n}^{11}\) system, meaning that all
\emph{V\textsubscript{n}} vectors have exactly one ``1'', one ``-1'',
and (\(n - 2)\) zeros. These vectors coincide with root system \(A_{n-1}\)
vectors~\cite{roots}. For NN training purposes
\emph{n} coincides with the number of LS dimensions.

\begin{figure}
	\centering
	\includegraphics[scale=0.45]{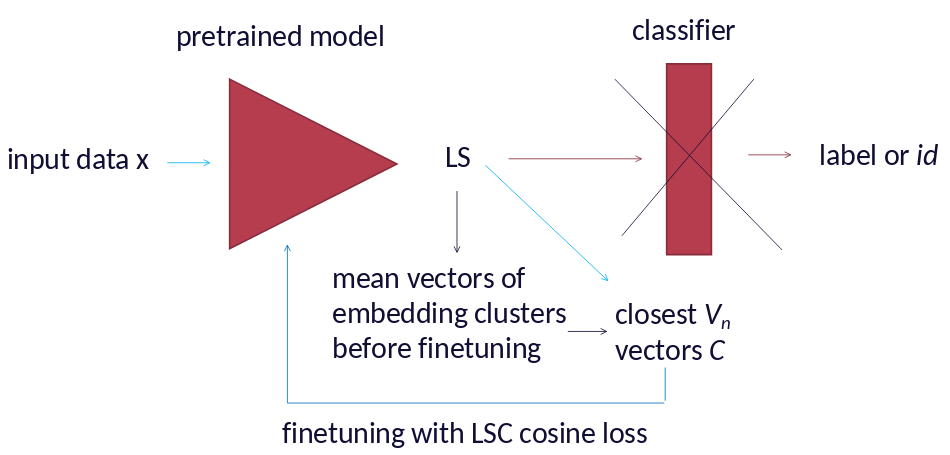} 
	\caption{Pretrained model finetuning with LSC. Black arrows
		indicate operations done once before finetuning. Blue arrows indicate
		the parameter update loop.}
	\label{fig:211}
\end{figure}
\unskip

\subsection{Finetuning LS with the original number of dimensions}
\label{finetune-norig}

Figure~\ref{fig:211} shows steps required to finetune a pretrained model using
LSC. Firstly, it is necessary to calculate class or \emph{id} mean
embedding vectors on the target dataset. A classification layer should
be removed if it is present in the architecture. Secondly, mean
\emph{id} embedding vectors should be paired with closest (with respect
to cosine similarity) unique vectors from the chosen
\emph{V\textsubscript{n}}. This allows to determine which \emph{id}
corresponds to which \emph{V\textsubscript{n}} vector and use them as
the target cluster center vector \emph{C}. Afterward the finetuning
process follows basic LSC training procedure described in Section 3.3 in~\cite{lsc2}. 
\emph{Id}s can be predicted in the same manner in which
labels are predicted in the original method, \emph{i.e.} by checking the
\emph{id} assigned to \emph{V\textsubscript{n}} vector closest to a
given embedding. \emph{Id} prediction accuracy used in Section~\ref{exp-finetune} is
the percentage of correct \emph{id} predictions for all data points in
the dataset used for finetuning.

\subsection{Finding V\textsubscript{n} vectors closest to mean embedding vectors}
\label{closevn}

It has been mentioned in Section~\ref{finetune-norig} that each \emph{id} should be
paired with the closest \emph{V\textsubscript{n}} vector. However, two
or more \emph{id} clusters may share the same closest
\emph{V\textsubscript{n}} vector, which can be expected for classes or
\emph{ids} of objects or persons with some degree of similarity.
Assigning the same \emph{V\textsubscript{n}} vector to several
\emph{id}s will make these \emph{id}s indistinguishable after
finetuning, which is not acceptable. Hence, all but one \emph{id} should
be assigned different \emph{V\textsubscript{n}} vectors.

Fortunately, there exists a neat way of finding possible
\emph{V\textsubscript{n}} vectors owing to the structure of the proposed
vector systems. It has previously been shown that for an arbitrary
embedding \emph{m} of its largest and \emph{k} of its smallest elements
determine its cosine similarity with \(V_{n}^{mk}\) vectors~\cite{lsc4}.
For \(V_{n}^{11}\) this means that only maximum and minimum magnitude
elements in the embedding affect the similarity calculation results, and
one can find their indexes using argmax and argmin operations,
respectively. It should be mentioned that \(V_{n}^{11}\) vectors can be
uniquely referred to using indexes of their positive and negative
elements. For example, a vector with elements {[}1,-1, 0 ... 0{]} is
referred to as (0, 1), another one with elements {[}0,1, 0 ... -1{]} is
referred to as (1, \emph{n}-1), etc. This pair is hereafter referred to
as index of \(V_{n}^{11}\) vector.

\begin{figure}
	\centering
	\includegraphics[scale=0.45]{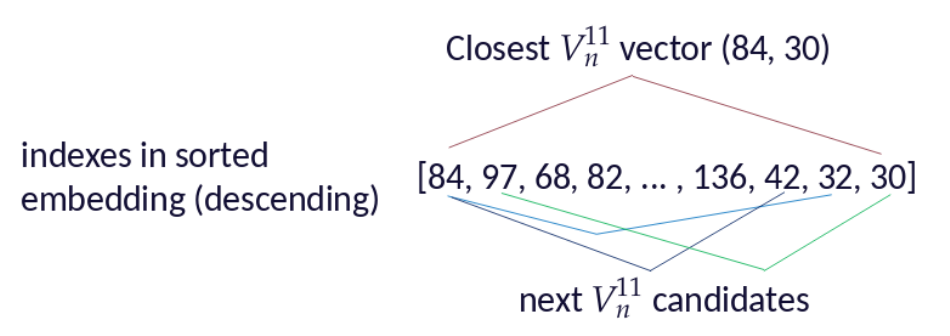} 
	\caption{Example of finding \(V_{n}^{11}\) vectors close to some
		mean embedding vector when the closest \(V_{n}^{11}\) vector is already
		assigned an \emph{id}.}
	\label{fig:221}
\end{figure}

Large and small elements of embedding vectors, as well as their indexes,
can also be found by analyzing sorted embedding vectors. When the
closest vector, \emph{i.e.,} a vector with ``1'' in embedding's argmax
position and ``-1'' in its argmin position, to some mean embedding
vector is already assigned an \emph{id,} another close
\emph{V\textsubscript{n}} vector can be found by choosing pairs of
indexes as shown in Figure~\ref{fig:221}. This approach ensures that mean
embedding vectors are not paired with distant \emph{V\textsubscript{n}}
vectors and the original distribution of clusters is not disturbed.

\unskip

\subsection{Finetuning LS with the reduced number of dimensions}
\label{finetune-nred}

Reducing the number of dimensions of NN embeddings can lead to faster
computations and storage space reduction~\cite{BB, ADR}. However, this
often leads to worse accuracy metrics, which indicates an important
trade-off. Hence, computation acceleration and accuracy metrics should
be carefully monitored to identify the optimal number of LS dimensions.

In this paper we use random projection to obtain low-dimensional NN
embeddings~\cite{proof, RP}. Specifically, to obtain embeddings in
LS with reduced number of dimensions \emph{n\textsubscript{red
}}(hereafter referred to as reduced LS) using known embeddings in LS
with the original number of dimensions \emph{n\textsubscript{orig}}, we
first construct a random uniform matrix with dimensions
{[}\emph{n\textsubscript{orig}}, \emph{n\textsubscript{red}}{]}. We then
apply \emph{QR} decomposition to this matrix~\cite{QR} and use the
obtained matrix \emph{Q} (with dimensions
{[}\emph{n\textsubscript{orig}}, \emph{n\textsubscript{red}}{]}) for
embedding projection

\begin{equation}
	z_{red} = z_{orig} \cdot Q,
	\label{eq:z_red}
\end{equation}
\unskip

where \emph{z\textsubscript{orig}} and \emph{z\textsubscript{red}} are
NN embeddings in the original and reduced LS, respectively. It has
previously been shown that for specific \emph{n\textsubscript{id}},
\emph{n\textsubscript{red}} should be chosen to be the smallest possible
when the number of vectors in \emph{V\textsubscript{n}} system
\emph{n\textsubscript{vects}} \(\geq\) \emph{n\textsubscript{id}}~\cite{lsc3}. 
In case of \(V_{n}^{11}\) the number of vectors is

\begin{equation}
	n_{vects} = n \cdot (n - 1).
	\label{eq:n_vects}
\end{equation}
\unskip

\begin{figure}
	\centering
	\includegraphics[scale=0.45]{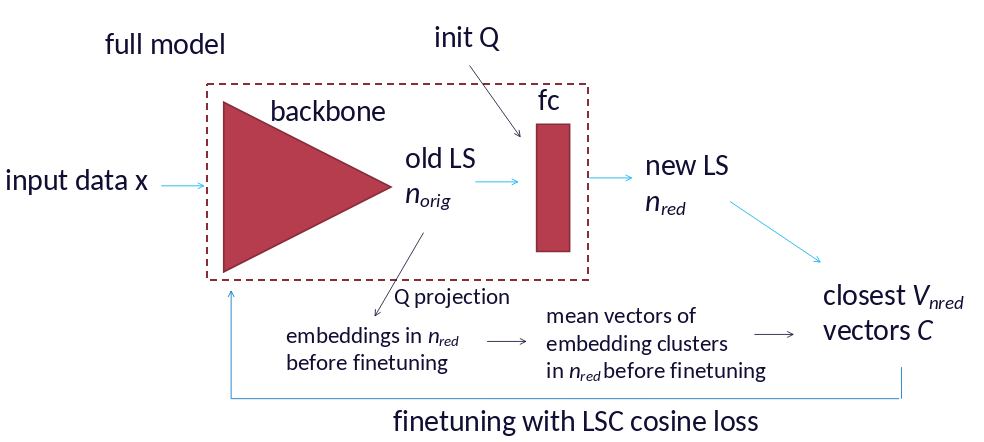} 
	\caption{Pretrained model finetuning while reducing the number of LS
		dimensions. Here \emph{fc} is fully-connected projection layer
		initialized with the same matrix \emph{Q}, which is used to project
		embeddings from \emph{n\textsubscript{orig}} into
		\emph{n\textsubscript{red}}. Black arrows indicate operations done once
		before finetuning. Blue arrows show the parameter update loop.}
	\label{fig:231}
\end{figure}

Substituting \emph{n\textsubscript{red}} for \emph{n},
\emph{n\textsubscript{id}} for \emph{n\textsubscript{vects}}, and
solving for \emph{n\textsubscript{red}} yields

\begin{equation}
	n_{red} = \frac{1 + \sqrt[]{\left( 1 + {4 \cdot n}_{id} \right)}}{2},
	\label{eq:z_red_fin}
\end{equation}
\unskip

which should be rounded up to the next integer. Figure~\ref{fig:231} shows that
the projected embeddings are used to calculate mean vectors of embedding
clusters and to find closest vectors in \emph{V\textsubscript{n}} to use
them as vector \emph{C} as discussed in Section~\ref{finetune-norig}. A projection layer
should also be included into NN architecture so output embeddings'
number of dimensions matches \emph{n\textsubscript{red}}. This
projection layer should be initialized with the same matrix \emph{Q}
which is used to project embeddings.

\unskip

\section{Search algorithms for fast retrieval of top \emph{K} relevant images}
\label{search}

In this paper we consider exhaustive $\mathcal{O}(n\textsubscript{g})$ search
as baseline for comparison. Algorithms proposed in this Section focus on
finding a small gallery subset for similarity calculation using
geometric properties of \emph{V\textsubscript{n}} used for LSC.
Specifically, we first find closest \emph{V\textsubscript{n}} vectors
for all gallery embeddings and store their indexes (as discussed in
Section~\ref{closevn}). Then for a query \emph{q} we find its closest
\emph{V\textsubscript{n}} vector, retrieve all gallery embeddings with
the same closest \emph{V\textsubscript{n}} vector (called candidates),
and then compare \emph{q} only with these candidates. Since the closest
\emph{V\textsubscript{n}} vector search complexity is $\mathcal{O}(1)$, the
complexity of the candidate subset search is also $\mathcal{O}(1)$, and the
overall complexity is $\mathcal{O}(n\textsubscript{c})$ rather than
$\mathcal{O}(n\textsubscript{g})$, so the speed gain is proportional to

\begin{equation}
	\frac{t_{exh}}{t_{new}}\ \sim\ \frac{n_{g}}{n_{c}(id)} \approx n_{id},
	\label{eq:speedup}
\end{equation}
\unskip

where \emph{t\textsubscript{exp}} and \emph{t\textsubscript{new}} are
time intervals required to retrieve top \emph{K} elements using the
exhaustive search and the proposed algorithms, respectively. Thus, the
more \emph{id}s the gallery has, the higher acceleration can be
expected.

\begin{algorithm}[t]
	\caption{Candidate subset search and similarity calculation in
		structured LS for fast top \emph{K} retrieval using a hash table of
		closest \emph{V\textsubscript{n}} vectors (PyTorch-style pseudocode)}
	\label{alg:1}
	\begin{algorithmic}[1]
		\Require Gallery of embeddings \emph{G}, batch of queries
			\emph{q}, \emph{K} most similar elements to retrieve\emph{,} parameters
			\emph{m}, \emph{k} of vector system \emph{V\textsubscript{n}}
		
		\Statex \textbf{Preliminary step}
		\State $G_{\text{m\_indexes}},\ G_{\text{k\_indexes}} \gets \text{closest\_Vn}(G, m, k)$ \Comment{see~\cite{lsc4} for details}
		\State $G_{\text{dict}} \gets \{(G_{\text{m\_index}},\ G_{\text{k\_index}}) : [\text{indexes of } z \text{ in } G \text{ with same } (G_{\text{m\_index}},\ G_{\text{k\_index}})]\}$
		
		\Statex \textbf{Search step}
		\State $q_{\text{m\_indexes}},\ q_{\text{k\_indexes}} \gets \text{closest\_Vn}(q, m, k)$
		\State $\text{candidates} \gets \text{tensor}(G[G_{\text{dict}}[(q_{\text{m\_indexes}},\ q_{\text{k\_indexes}})])$
		\State $\text{similarities} \gets \text{cosine\_similarity}(q, \text{candidates})$
		\State $\_,\ \text{max\_indexes} \gets \text{topk}(\text{similarities}, k=K)$ \Comment{indexes of top $K$ in candidate subset}
		\State \textbf{return} $G[\text{candidates}[\text{max\_indexes}]]$ \Comment{indexes of top $K$ in $G$}
	\end{algorithmic}
\end{algorithm}

\subsection{CPU-focused search acceleration}
\label{search-cpu}

In this paper we present two possible implementations of the proposed
search method. Algorithm~\ref{alg:1} uses the original fast search method~\cite{lsc4}.
Here a hash table (\emph{e.g.}, a python dictionary) with
indexes of closest \emph{V\textsubscript{n}} vectors to all embeddings
in \emph{G} as keys and lists of all embeddings with the same closest
\emph{V\textsubscript{n}} vectors as values is created. The number of
keys in this hash table is proportional to \emph{n\textsubscript{id}},
and the number of values in each list is proportional to
\emph{n\textsubscript{c}}. For each query its closest
\emph{V\textsubscript{n}} vector is used to retrieve candidates using
this dictionary, which allows to form candidate tensor and compute
similarity. This search is fast due to hash table $\mathcal{O}(1)$ key search
complexity.

There are two main problems with this implementation when queries are
processed in batches. Firstly, dictionary search and data preparation do
not efficiently work in parallel. Secondly, since the number of
candidates varies for different queries in the batch, padding is
required when forming the candidate tensor. This means that candidate
tensor shape is always determined by the largest
\emph{n\textsubscript{c}} in the batch, which is suboptimal.
Nevertheless, Algorithm~\ref{alg:1} performs extremely well on CPU, as will be
shown in Section~\ref{Exp}.

\subsection{GPU-focused search acceleration}
\label{search-gpu}

Algorithm~\ref{alg:2} addresses the discussed issues of Algorithm~\ref{alg:1} by abandoning
the hash table and instead calculating and storing the closest
\emph{V\textsubscript{n}} vector indexes of gallery embeddings as a
tensor. When closest \emph{V\textsubscript{n}} vector indexes are
calculated for queries, boolean matching can be performed directly on
tensors of indexes. Two
vectors with indexes of candidates in \emph{G} and corresponding indexes
of queries in the batch are formed after finding nonzero elements. 
These indexes are used to retrieve
relevant embeddings and to compute similarity between each query and its
candidates as a single vector, so padding due to the varying
\emph{n\textsubscript{c}} is not needed. As the cosine similarity is
bounded to {[}-1, 1{]} region, doubled query index is substracted from the
similarity values to prevent the results corresponding to different
queries in the batch from mixing. Finally, the result is sorted according
to shifted similarity, split using query indexes, and the
first \emph{K} elements for every query are returned.

It should be noted that in this work the candidates are selected as
logical AND of matches between indexes corresponding to \emph{m=1}
largest and \emph{k=1} smallest elements in embedding vectors. This
algorithm can be extended to other vector systems simply by amending
this statement so it matches the required \emph{m} and \emph{k}
combination. The main advantage of Algorithm~\ref{alg:2} is that all its
operations can be implemented using PyTorch functions and executed in
parallel on the GPU which leads to significant overall acceleration, as
will be shown in Section~\ref{Exp}. The only drawback is that the matching
operation uses gallery-size objects, so Algorithm~\ref{alg:2} is slower than
Algorithm~\ref{alg:1} on CPU.

\begin{algorithm}
	\caption{Candidate subset search and similarity calculation in
		structured LS for fast top \emph{K} retrieval using tensor operations (PyTorch-style pseudocode)}
	\label{alg:2}
	\begin{algorithmic}[1]
		\Require Gallery of embeddings $G$, batch of queries
			$q$, $K$ most similar elements to retrieve, parameters
			$m$, $k$ of vector system $V_{\text{n}}$
		
		\Statex \textbf{Preliminary step}
		\State $G_{\text{m\_indexes}},\ G_{\text{k\_indexes}} \gets \text{closest\_Vn}(G, m, k)$ \Comment{as torch.tensor}
		
		\Statex \textbf{Search step}
		\State $q_{\text{m\_indexes}},\ q_{\text{k\_indexes}} \gets \text{closest\_Vn}(q, m, k)$ \Comment{as torch.tensor}
		\State $\text{matches} \gets (q_{\text{m\_indexes}}[:, \text{None}] == G_{\text{m\_indexes}}[\text{None}, :])\ \&$
		\Statex $\phantom{\text{matches} \gets}\ q_{\text{k\_indexes}}[:, \text{None}] == G_{\text{k\_indexes}}[\text{None}, :])$
		\State $q_{\text{matches}},\ G_{\text{matches}} \gets \text{matches.nonzero}(\text{as\_tuple=True})$
		\State $\text{similarities} \gets \text{cosine\_similarity}(q[q_{\text{matches}}], G[G_{\text{matches}}])$
		\State $\text{similarities} \gets \text{similarities} - 2 \cdot q_{\text{matches}}$
		\State $\_,\ \text{indexes} = \text{torch.sort}(\text{similarities}, \text{descending=True})$
		\State $\_,\ \text{counts} = \text{torch.unique\_consecutive}(q[q_{\text{matches}}], \text{return\_counts=True})$
		\State $\text{real\_indexes} = \text{indexes}[\text{counts}]$
		\State \textbf{return} $G[G_{\text{matches}}][\text{real\_indexes}]$ \Comment{indexes of top $K$ in $G$}
	\end{algorithmic}
\end{algorithm}

\section{Experiments}
\label{Exp}

In this study the effectiveness of the retrieval operation is measured
using rank-\emph{N} (or RN) metric, which indicates whether a relevant
(correct \emph{id}) element is among \emph{N} highest similarity (or top
\emph{K}, where \emph{K}=\emph{N}) gallery images~\cite{MegaFace,
ArcFace}. We use IResNet50~\cite{ArcFace} pre-trained with ArcFace
algorithm on MS-Celeb-1M~\cite{msc} from InsightFace library~\cite{InsightFace}
as the backbone model. The model is finetuned on
target dataset to configure its LS, and then rank-1 (R1) and rank-10
(R10) metrics are calculated using exhaustive search and the algorithms
proposed in Section~\ref{search}. \(V_{n}^{11}\) vector system is used for LS
configuration in all finetuning experiments. The original LS dimension
of IResNet50 \emph{n\textsubscript{orig}}=512, and the reduced LS
dimension found using~(\ref{eq:z_red_fin}) is \emph{n\textsubscript{red}}=160.

The target dataset is a subset of MS-Celeb-1M with approximately 1.48m
images and 24.3k \emph{ids}. Since gallery/query splits are not provided
for this dataset, 10\% of images of every \emph{id} are randomly chosen
as queries, while the rest form the gallery. These splits are fixed for
all experiments. Standard PyTorch dataloaders with batch size of 256 are
used for image loading. Models are finetuned for 50 epochs using AdamW
optimizer with 10\textsuperscript{-5} learning rate and
10\textsuperscript{-5} weight decay~\cite{Adam, AdamW}. Batch size for
rank-\emph{N} calculation is 64 (reduced because full gallery needs to
be stored on GPU), and we report the total time required to calculate
rank-\emph{N} metrics on \emph{G} for all queries. All experiments are
conducted using Intel(R) Xeon(R) Gold 6230R CPU and a single 40GB NVIDIA
A100 GPU. All PyTorch CPU operations are limited to two CPU threads.

\begin{figure}
	\centering
	\includegraphics[scale=0.5]{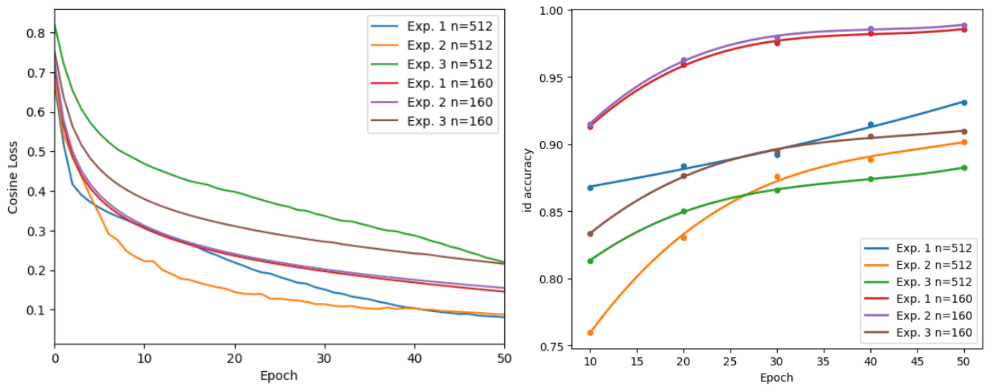} 
	\caption{Curves of loss (left) and \emph{id} prediction accuracy
		(right) during finetuning.}
	\label{fig:411}
\end{figure}
\unskip

\subsection{Finetuning experiments}
\label{exp-finetune}

IResNet50 models are finetuned using the methodology described in
Section~\ref{lscfinetune}. Metrics of the pretrained model on the target dataset before
finetuning are shown in Tables~\ref{tab:421} and~\ref{tab:422}. It should be noted that
since this model is pretrained on the complete MS-Celeb-1M, its training
dataset includes both query and gallery images. However, in realistic
application scenarios query images come from live camera feeds and are
not available during training. This discrepancy is addressed in our
finetuning setups as discussed below.

Three finetuning scenarios are considered in this study. Similar to
pretraining, the complete dataset is used for finetuing in Experiment I.
On the contrary, only gallery images are used for finetuing in
Experiment II. Queries are also not used for finetuning in Experiment
III, and a set of light augmentations is applied to gallery images to
improve generalization. Hence, experiments II and III represent test
scenarios better.

Finetuning loss values and \emph{id} accuracy (calculated every 10
epochs starting at 10\textsuperscript{th} epoch) curves shown in Figure~\ref{fig:411}
illustrate that neither loss nor accuracy have plateaued at
50\textsuperscript{th} epoch. This indicates that the metrics (including
rank-\emph{N} metrics) can be further improved by increasing the number
of epochs and better hyperparameter tuning.

\subsection{Rank-N metrics and calculation time}
\label{exp-rankn}

Tables~\ref{tab:421} and~\ref{tab:422} show that Algorithms~\ref{alg:1} and~\ref{alg:2} are consistently
faster than the exhaustive search on GPU. Specifically, Algorithm~\ref{alg:2} on
GPU is almost 3 times faster, while Algorithm~\ref{alg:1} on CPU is about 40\%
faster. The latter is remarkable since a CPU method happens to be faster
than a GPU one. Both exhaustive search and Algorithm~\ref{alg:2} on CPU are
significantly slower than Algorithm~\ref{alg:1}. The average sizes of candidate
sets used for similarity calculation in Algorithms~\ref{alg:1} and~\ref{alg:2} are about 100
embeddings, which is the main reason for the reported acceleration.

\begin{table}
	\caption{Gallery ranking metrics and time required for their
		calculation for \emph{n\textsubscript{orig}}=512.} 
	\label{tab:421}
	\centering
	\begin{tabular}{|c|c|c|c|c|c|c|}
	  \hline
		Method & Device & Time, s & Pretrained & Exp I & Exp II & Exp III \\
		& & & R1/R10 & R1/R10 & R1/R10 & R1/R10 \\ \hline
		Queries in train set & - & - & Yes & Yes & No & No \\ \hline
		Augmentation & - & - & Yes & No & No & Yes \\ \hline
		exh. search & GPU & 17.5 & 0.884/0.928 & 0.944/0.964 & 0.801/0.83 &
		0.83/0.86 \\ \hline
		exh. search & CPU & 336.4 & 0.884/0.928 & 0.944/0.964 & 0.801/0.83 &
		0.83/0.86 \\ \hline
		Alg. 1 & CPU & 12.5 & 0.266/0.27 & 0.942/0.958 & 0.798/0.81 &
		0.818/0.835 \\ \hline
		Alg. 2 & GPU & 5.6 & 0.266/0.27 & 0.942/0.958 & 0.798/0.81 &
		0.818/0.835 \\ \hline
		Alg. 2 & CPU & 84.1 & 0.266/0.27 & 0.942/0.958 & 0.798/0.81 &
		0.818/0.835 \\ 
		\hline
	\end{tabular}
\end{table}

\begin{table}
	\caption{Gallery ranking metrics and time required for their
		calculation for \emph{n\textsubscript{red}}=160.} 
	\label{tab:422}
	\centering
	\begin{tabular}{|c|c|c|c|c|c|c|}
	  \hline
		Method & Device & Time, s & Pretrained & Exp I & Exp II & Exp III \\
		& & & R1/R10 & R1/R10 & R1/R10 & R1/R10 \\ \hline
		Queries in train set & - & - & Yes & Yes & No & No \\ \hline
		Augmentation & - & - & Yes & No & No & Yes \\ \hline
		exh. search & GPU & 12.9 & 0.87/0.914 & 0.97/0.99 & 0.82/0.854 &
		0.846/0.875 \\ \hline
		exh. search & CPU & 280 & 0.87/0.914 & 0.97/0.99 & 0.82/0.854 &
		0.846/0.875 \\ \hline
		Alg. 1 & CPU & 7.1 & 0.266/0.27 & 0.97/0.99 & 0.786/0.796 &
		0.825/0.839 \\ \hline
		Alg. 2 & GPU & 4.3 & 0.266/0.27 & 0.97/0.99 & 0.786/0.796 &
		0.825/0.839 \\ \hline
		Alg. 2 & CPU & 75.2 & 0.266/0.27 & 0.97/0.99 & 0.786/0.796 &
		0.825/0.839 \\ \hline
	\end{tabular}
\end{table}

\section{Discussions}
\label{discussions}

\subsection{LSC finetuning, ranking metrics, and dimensionality reduction}
\label{disc-finetuning}

This paper describes how LSC can be used to finetune pretrained NN
models for downstream tasks. In this case the preexisting cluster
distribution formed during pretraining is restructured during
finetuning. The existence of non-trivial starting cluster distribution
allows to assign labels or \emph{id}s to \emph{V\textsubscript{n}}
vectors in a consistent manner solving the labeling ambiguity problem
which exists when models are trained with LSC from scratch~\cite{lsc2,lsc3}. 
Hence, the distribution of \emph{id}s among
\emph{V\textsubscript{n}} vectors is determined by the cluster
distribution formed by NN during pretraining, and the final LS geometry
is determined by \emph{V\textsubscript{n}} during finetuning.

The comparison of the exhaustive search results between the pretrained
model and Experiment~1 (queries are included in training
sets in both cases) in Tables~\ref{tab:421} and~\ref{tab:422} shows that finetuning can
improve rank-\emph{N} metrics. As might be expected, removing queries
from the training dataset leads to reduction in metrics. As mentioned in
Section~\ref{exp-finetune}, metrics can be improved by increasing the number of epochs
and hyperparameter tuning. However, it should be kept in mind that the
main goal of the LSC finetuning is to form structures in LS so that
Algorithms~\ref{alg:1} and~\ref{alg:2}, which rely on specific properties of
\emph{V\textsubscript{n}} geometry, do not reduce metrics when compared
to the exhaustive search results. Tables~\ref{tab:421} and~\ref{tab:422} show that
rank-\emph{N} metrics calculated using Algorithms~\ref{alg:1} and~\ref{alg:2} before
finetuning are extremely low. This happens since the embedding cluster
distribution of pretrained model is very different from the
\emph{V\textsubscript{n}} vector distribution. This illustrates that the
proposed algorithms cannot be used for NNs which LS is not structured.
On the contrary, metrics calculated using Algorithms~\ref{alg:1} and~\ref{alg:2} after
finetuning are a close match to the metrics calculated using the
exhaustive search, showing that the required LS structure has indeed
been achieved.

It should be mentioned that sometimes for tricky queries one might want
to expand the search. In case of exhaustive search this is achieved by
increasing \emph{K} or \emph{N}, since similarities are always
calculated with the whole gallery. However, increasing \emph{K} might
not give expected results in the proposed methods when
\emph{n\textsubscript{c}} is small, because similarities are calculated
only with embeddings that share the same closest
\emph{V\textsubscript{n}} vectors with the query. Hence, for
\(K > n_{c}\) one needs to also consider embeddings which are in
vicinity of other \emph{V\textsubscript{n}} vectors that are close to
the query embedding (as described in Section~\ref{closevn}). This extension to the
method will be studied in greater detail in the future.

Tables~\ref{alg:1} and~\ref{alg:2} also show that reducing the number of LS
dimensions leads to faster computations. Moreover, it can also result in
better metrics when NN is finetuned directly in
\emph{n\textsubscript{red}}. Whereas the reduction in the number of LS
dimensions is about 300\% in this work, this number can be much higher
when vector systems with larger \emph{n\textsubscript{vects}} are used~\cite{lsc3}.
Potential acceleration due to dimensionality reduction is
extremely promising when NN models are finetuned on datasets with known
\emph{n\textsubscript{id}}.

\subsection{Application scenarios of the proposed algorithms}
\label{disc-applic}

The results in Tables~\ref{alg:1} and~\ref{alg:2} show two distinct application
scenarios of the proposed method. Firstly, Algorithm~1 can be used when
computational budget is low and GPU usage is limited. This is extremely
relevant for edge devices where computationally efficient methods are
essential~\cite{Edge, NRaa}. Algorithm~\ref{alg:1} can also be used in traditional
high performance settings to reduce GPU usage of search operations and
to divert computations to other elements of the pipeline.

Secondly, Algorithm~\ref{alg:2} can substitute conventional GPU search methods in
systems with huge galleries and extremely large
\emph{n\textsubscript{id}}. It can provide significant acceleration with
minimal pipeline modifications while simultaneously reducing required
memory due to the absence of large matrix multiplication for similarity
evaluation. Acceleration and storage memory reduction are also achieved
when NN is finetuned in \emph{n\textsubscript{red}}. Combining the
benefits of the proposed algorithms with advantages achieved by LS
dimensionality reduction can lead to significant speed-up and reduction
in total memory required to process and store query and gallery
embeddings, which is essential for efficient operation of modern
systems.

\section{Conclusions}
\label{conclusions}

This paper describes LSC finetuning methodology and fast ranking
algorithms which rely on geometric properties of vector systems used to
configure LS of NNs. Application of these algorithms leads to up to
300\% speed-up on GPU and up to 40\% speed-up on CPU compared to
baseline exhaustive search on GPU. It is shown that retrieval metrics
are very low when these algorithms are applied to embeddings of
pretrained models with unstructured LS. On the contrary, the metrics
calculated after finetuning almost match the exhaustive search metrics,
which indicates that LS geometry matches the target vector system. It is
also shown that finetuning in LS with the reduced number of dimensions
allows faster operation and required memory reduction while increasing
retrieval metrics. The proposed approaches are promising for
industry-level applications as well as for usage in low performance
systems and edge devices.

\section*{Acknowledgement}
\label{acknowledgement}

The authors would like to thank Ilya Androsov for suggesting random
projection with \emph{Q} matrix for dimensionality reduction, Dr Igor
Netay for discussions, and Vasily Dolmatov for fruitful discussions and
project supervision.


\bibliographystyle{IEEEtran}
\bibliography{IEEEabrv,ms}

\end{document}